%% file: main.tex
\documentclass[10pt,twocolumn,letterpaper]{article}

\usepackage{cvpr}              

\usepackage[utf8]{inputenc}
\usepackage{tcolorbox}
\usepackage{tabularx}
\usepackage{graphicx}
\usepackage{makecell}

\tcbset{
  colback=blue!5,     
  colframe=blue!70!black, 
  boxrule=1.5pt,      
  arc=3mm,            
  left=6pt, right=6pt, top=6pt, bottom=6pt 
}

\definecolor{cvprblue}{rgb}{0.21,0.49,0.74}
\usepackage[pagebackref,breaklinks,colorlinks,allcolors=cvprblue]{hyperref}

\def\paperID{} 
\def\confName{CVPR}
\def\confYear{2026}

\title{Pseudo-Unification: Entropy Probing Reveals Divergent Information Patterns in Unified Multimodal Models}

\author{Songlin Yang, Xianghao Kong, Anyi Rao\\
MMLab@HKUST, The Hong Kong University of Science and Technology\\
{\tt\small syangds@connect.ust.hk}
}

\begin{document}
\maketitle
\input{sec/0_abstract}    
\input{sec/1_intro}

\input{sec/2_related_work}

\input{sec/3_theoretical_probing}

\input{sec/4_probing_framework_setting}

\input{sec/5_prompt}

\input{sec/6_response}

\input{sec/7_conclusion}
{
    \small
    \bibliographystyle{ieeenat_fullname}
    \bibliography{main}
}


\end{document}

%% file: sec/0_abstract.tex
\begin{abstract}

Unified multimodal models (UMMs) were designed to synergize the creative reasoning of large language models (LLMs) with the fidelity-driven generation of vision models. In practice, however, this synergy remains elusive: UMMs fail to transfer LLM-like reasoning to image synthesis, exhibiting divergent response behaviors. We term this phenomenon \textbf{pseudo-unification}. Diagnosing its model-internal causes is crucial, but existing probing methods either lack model-internal insight or ignore prompt–response dependencies. To address these probing limitations, we propose an information-theoretic probing framework that jointly analyzes how UMMs encode inputs and generate outputs. Applied to ten representative UMMs, our framework reveals that pseudo-unification stems from a dual divergence: (i) Modality-Asymmetric Encoding, where vision and language follow divergent entropy trajectories, and (ii) Pattern-Split Response, where text generation exhibits high-entropy creativity while image synthesis enforces low-entropy fidelity. Only models that unify both sides (e.g., via contextual prediction) achieve more genuine unification, enabling stronger reasoning-based text-to-image generation even with fewer parameters. Our work provides the first model-internal probing of unification, demonstrating that real multimodal synergy requires consistency in information flow, not just shared parameters.

\end{abstract}

%% file: sec/1_intro.tex
\section{Introduction}
\label{sec:intro}

\begin{figure}[t]
    \centering
    \includegraphics[width=\linewidth]{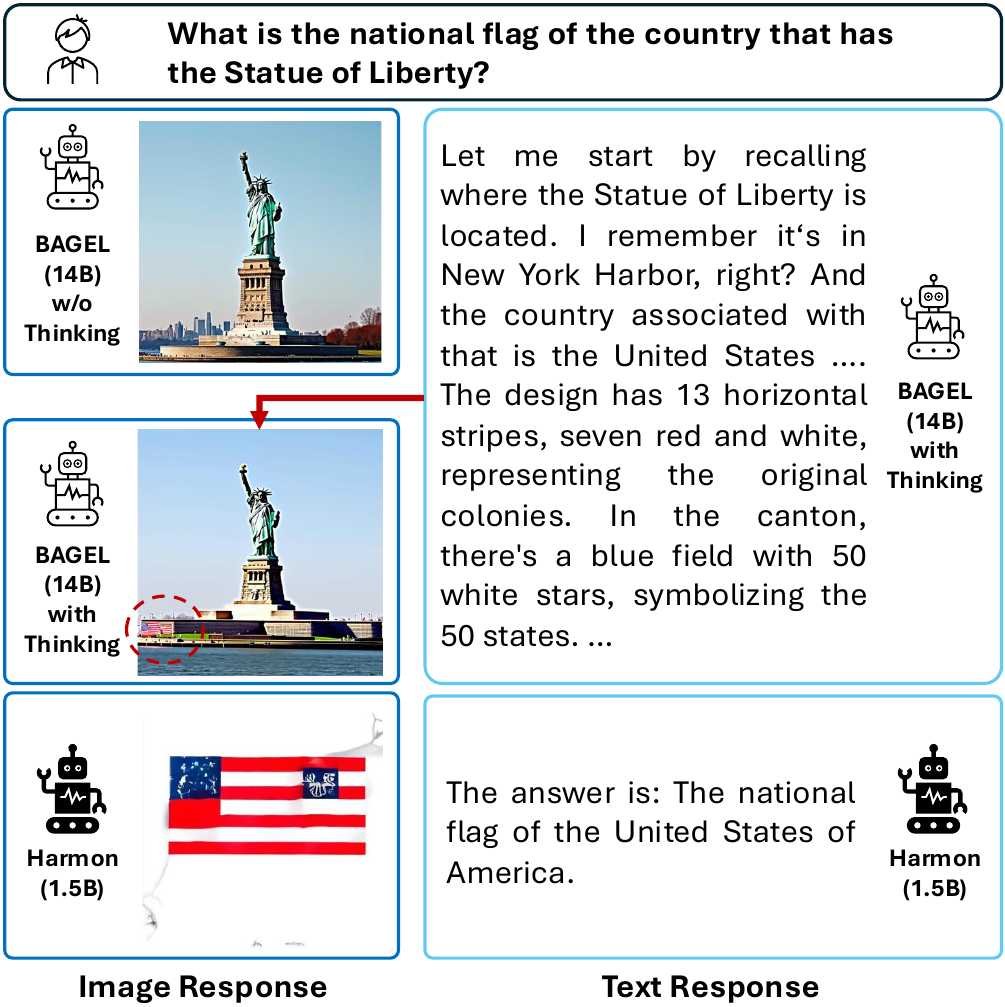}
    \vspace{-0.4cm}
    \caption{\textbf{An Illustration of \textit{Pseudo-Unification}.} We conduct an ``unfair'' comparison between BAGEL (14B)~\citep{bagel} and the much smaller Harmon (1.5B)~\citep{harmon} on a reasoning task about the American flag. Two key observations emerge: (i) \textit{Response Divergence}: text correctly retrieves “American flag,” but image generation fails to produce it; (ii) \textit{Superior Cross-Modal Reasoning in a Small Model}: despite lower fidelity and shorter outputs, Harmon aligns both modalities around the core concept.}
    \label{fig:teaser}
    \vspace{-0.5cm}
\end{figure}

Before unified multimodal models (UMMs)~\citep{harmon,bagel,emu3,yang2026shotverse,tang2026endogenous,li2025instant}, text generation and image synthesis were optimized under distinct objectives. Large language models (LLMs)~\citep{gpt-4,deepseek_,qwen} learned a creative response pattern through next-token prediction, emphasizing contextual plausibility over strict input alignment. In contrast, text-to-image (T2I) models~\citep{rombach2022high,flux2024,li2025beyond,yang2026human,lialpha} trained on text–image pairs favored fidelity to the prompt. UMMs were expected to unify these paradigms and enable cross-modal synergy. Yet, as illustrated in Fig.~\ref{fig:teaser}, the creative, reasoning-based generation ability of LLMs has not transferred to image synthesis: UMMs still struggle to interpret prompts, understand context, retrieve knowledge, and then generate images accordingly. Although tasks share a common representation space, their response patterns remain divergent. We term this phenomenon \textbf{\textit{pseudo-unification}}. Furthermore, as shown in Fig.~\ref{fig:teaser}, smaller models unexpectedly exhibit superior reasoning-based image generation performance. This discrepancy between expectation and practice of UMMs, coupled with their high training cost and rapid proliferation of architectures, underscores the need for unification probing.

While the community widely recognizes the importance of probing unification, current research still faces fundamental limitations: (i) Data-Driven Probing Lacks Model-Internal Insight: task-specific datasets~\citep{T2I-CoReBench,unieval,mme-unify} fail to capture the synergy, and new synergy benchmarks~\citep{zou2025uni,shi2025realunify}, still constrained to case studies, cannot reveal why certain models internally unify better than others; (ii) Current Model-Internal Analysis Overlooks Prompt-Response Dependencies: although LLM studies~\citep{cai2021isotropy,other-layer-by-layer,skean2023dime} have examined prompt representations layer by layer, they rarely investigate the prompt–response dependencies. Therefore, a more general probing framework is required, which checks internal information flow for diagnosing unification.

To address these probing limitations, we propose an information-theoretic probing framework that captures how UMMs encode (prompt) and generate (response) information across modalities. Our framework consists of two complementary levels: (i) Prompt Level: we analyze how text and image inputs are represented internally by comparing their embedding entropy and layer-wise entropy of hidden states. This reveals modality-specific differences in information preservation and compression. For instance, whether the visual stream reaches a representational bottleneck earlier than the linguistic one. (ii) Response Level: we probe their response behavior by estimating the conditional entropy between prompt and response representations across layers. The evolution of this conditional entropy reflects whether the model maintains consistent reasoning across modalities or exhibits divergent response patterns.

However, applying classical entropy estimation in information theory to UMMs is fundamentally infeasible. Transformer-based models~\citep{vaswani2017attention} do not expose explicit joint densities, and their representations (which are high-dimensional and variable-length) defy assumptions of density-based entropy estimation. This creates a critical theoretical gap: \textit{how can we quantify information flow in implicit spaces without access to underlying distributions?}

We bridge this gap through a reformulation of information measures in reproducing kernel Hilbert spaces (RKHS). Treating embedding sequences as empirical samples, we model their similarity via Gaussian kernels and reinterpret entropy not as a function of probability density, but as a geometric property of representation structure: Prompt entropy captures the intrinsic uncertainty (\textit{e.g.}, isotropy and spread) of a modality’s embeddings; Prompt-response joint entropy quantifies the structural richness; Their difference defines a non-parametric conditional entropy proxy, which measures the residual output uncertainty given the input. This reformulation is not merely a computational workaround. It provides a new operational semantics for information in deep implicit models. The proposed proxy behaves consistently with classical conditional entropy (\textit{i.e.}, low for faithful mappings, high for random ones), yet requires no density estimation.

After probing, we uncover that pseudo-unification arises from a dual divergence: (i) Prompt Representations are Modality-Asymmetric: vision and language follow divergent entropy trajectories, shaped by architectural priors rather than semantic content; (ii) Response Generation is Pattern-Split: text follows a  high-entropy creative pattern while image synthesis adheres to a low-entropy fidelity regime. Only models that align both encoding and generative logic (\textit{e.g.}, via contextual prediction) bridge this gap, revealing that real unification requires consistency not just in shared parameters, but in information flow. 


\textbf{\textit{Our work makes three key contributions}}:
(i) To investigate the internal causes of pseudo-unification in UMMs, we propose a two-level probing framework that disentangles unification into encoding consistency (via prompt entropy) and response coherence (via prompt–response conditional entropy), providing the first UMM diagnostic regarding model-internal information flow. (ii) To enable entropy estimation in Transformers, we reformulate information-theoretic measures in RKHS, allowing non-parametric computation of entropy and conditional entropy for implicit, variable-length representations. (iii) Through extensive analysis across ten representative UMMs, we show that pseudo-unification arises from a dual divergence: modality-asymmetric encoding and pattern-split responses, highlighting the model-internal causes of this phenomenon.

\begin{figure*}[t]
    \centering
    \includegraphics[width=\linewidth]{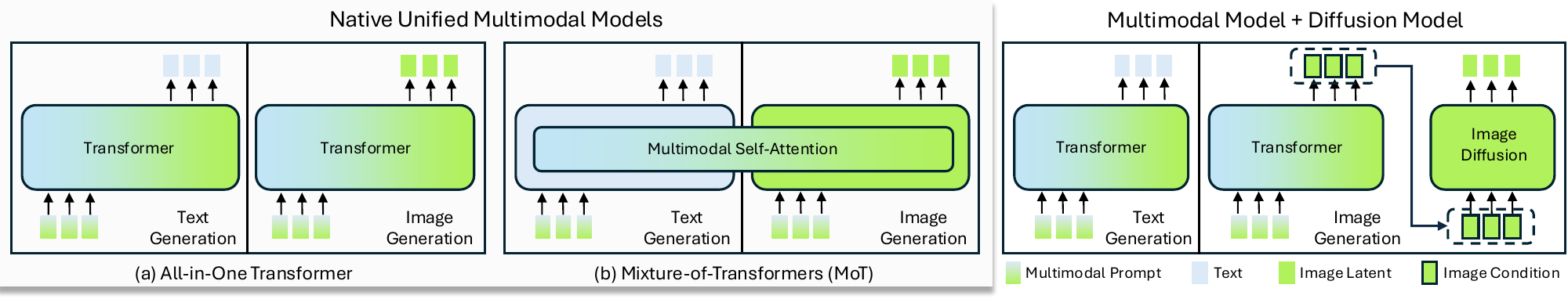}
    \vspace{-0.7cm}
    \caption{\textbf{Architectural Taxonomy of UMMs.} Current UMMs fall into two categories: (i) Native UMMs, which unify text and image generation within a single architecture (\textit{e.g.}, Harmon~\citep{harmon}, Janus-Pro~\citep{janus-pro}, and Show-o2~\citep{show-o2}), which employ an all-in-one Transformer to jointly produce text and image tokens, while BAGEL~\citep{bagel} uses a Mixture-of-Transformers (MoT)~\citep{mot} to separately generate text tokens and image tokens, fused via multimodal self-attention. (ii) MLLM + Diffusion Model Pipelines (\textit{e.g.}, OmniGen2~\citep{omnigen2}), where a multimodal LLM generates a text-based condition, and image synthesis is delegated to a separate diffusion model.}
    \label{fig:model_difference}
    \vspace{-0.3cm}
\end{figure*}

%% file: sec/2_related_work.tex
\section{Related Work}
\label{sec:related work}

\subsection{Unified Multimodal Models}


Unified multimodal models~\citep{emu3,janus-pro,chameleon,bagel,omnigen2,ovis-u1,uniworld-v1,Liquid,UniFluid,doracycle,blip3o} have emerged as a central research direction in multimodal intelligence. As shown in Fig.~\ref{fig:model_difference}, these diverse frameworks integrate both understanding and generation and have demonstrated competitive performance.

\noindent
\textbf{Evaluation for UMMs.} MME-Unify~\citep{mme-unify} jointly evaluate comprehension, generation, and mixed-modality tasks, and UniEval~\citep{unieval} operates without auxiliary models or human annotations. Complementary T2I benchmarks such as MMMG~\citep{MMMG}, T2I-CoReBench~\citep{T2I-CoReBench}, and WISE~\citep{wise}, can be adapted for UMMs but focus narrowly on image generation. Critically, none of these assess whether unifying understanding and generation yields synergistic gains.

\noindent
\textbf{Research and Evaluation for UMM Synergy.} Recent efforts~\cite{yan2025can} probe whether UMMs achieve real synergy between understanding and generation. On the training side, PairUni~\citep{zheng2025pairuni} aligns heterogeneous data into understanding–generation pairs and uses pair-aware policy optimization to reduce interference, while Co-Reinforcement Learning (CoRL)~\citep{jiang2025co} jointly optimizes both capabilities through unified RL stages to foster mutual improvement. Architecturally, Corvid~\citep{jiang2025corvid} employs hybrid visual encoders, cross-modal connectors, and inference-time chain-of-thought to enhance interpretable synergy. Meanwhile, benchmarks like RealUnify~\citep{shi2025realunify} and Uni-MMMU~\citep{zou2025uni} introduce bidirectionally coupled tasks and stepwise protocols to explicitly evaluate the two synergy axes, which are ``understanding enhances generation'' and ``generation enhances understanding'', revealing that co-locating capabilities in one model does not ensure effective cross-capability reinforcement. \textit{However, current probing and evaluation have not yet addressed the underlying model-internal causes of synergy, or its absence in UMMs.}

\subsection{Neural Representations in Language Models}

Regarding the semantic geometry of neural representations in language models, early efforts used linear probes \citep{alain2016understanding} and similarity metrics like SVCCA \citep{raghu2017svcca}, though many focused on vision or shallow networks. More recent studies examine which transformer layers encode linguistic or semantic structures, often identifying mid-depth layers as optimal for abstraction \citep{liu2019linguistic, tenney2019bert, voita2019bottom, jin2024conceptdepth, gurnee2023language, fan2024notalllayers}. Complementary theoretical analyses link pre-training objectives to representational structure \citep{saponati2025underlying}, while others explore phenomena like attention sinks \citep{attention-sinks, identifiability, gu2024attention, first-token-attending} and layer-wise compression–generalization trade-offs \citep{bordes2022guillotine, park2024geometry, deletanglanguage}. These studies further propose diverse metrics to quantify representation quality: information-theoretic measures (\textit{e.g.}, Information Bottleneck \citep{shwartz2017opening, shwartz2022information}, intrinsic dimensionality \citep{doimo-abstraction-phase, doimo-hidden-representations, anisotropy}), geometric properties (\textit{e.g.}, effective rank \citep{garrido2023rankme}, anisotropy \citep{anisotropy}, curvature \citep{hosseini2024curvature}), and task-based or invariance metrics (\textit{e.g.}, InfoNCE \citep{oord2018representation}, LiDAR \citep{thilak2023lidar}, NESum \citep{agrawal2022alphareq}).

\textit{However, this line of research is confined to language models and analyzes only prompt representations, ignoring prompt–response dependencies and multimodal joint reasoning. Our work bridges this gap by probing both input representations and prompt–response dynamics in UMMs, advancing from semantic geometry to a mechanistic understanding of unification.}

%% file: sec/3_theoretical_probing.tex
\section{An Entropy-Probing Formulation for Unification Analysis in UMMs}
\label{sec:entropy_formulation}

To model-internally diagnose the phenomenon of pseudo-unification in UMMs, we formalize unification as the learning of an implicit joint distribution over vision and language (Sec.~\ref{subsec:implicit_joint}). Within this formulation, the entropy of prompt representations and the conditional entropy of responses given prompts emerge as natural measures of prompt quality and response patterns, respectively (Sec.~\ref{subsec:info_theory}). However, because UMMs lack explicit probability densities and operate on variable-length embeddings, classic information-theoretic entropy estimation is infeasible (Sec.~\ref{subsec:challenges}).

Therefore, we first leverage matrix-based Rényi entropy (Sec.~\ref{subsec:matrix_entropy}) to quantify the intrinsic uncertainty and isotropy of embedding sequences in a non-parametric, kernel-based manner. This formulation enables direct comparison of representational structures across heterogeneous modalities and reveals how different UMMs encode inputs with varying degrees of information preservation and compression. Building on this foundation, we propose conditional entropy proxy (Sec.~\ref{subsec:proxy}) by measuring the additional structural complexity introduced by the response relative to the prompt. The resulting proxy provides a probe of response patterns, allowing us to diagnose whether a UMM truly unifies generative behavior across modalities.


\begin{figure*}[t]
    \centering
    \includegraphics[width=\linewidth]{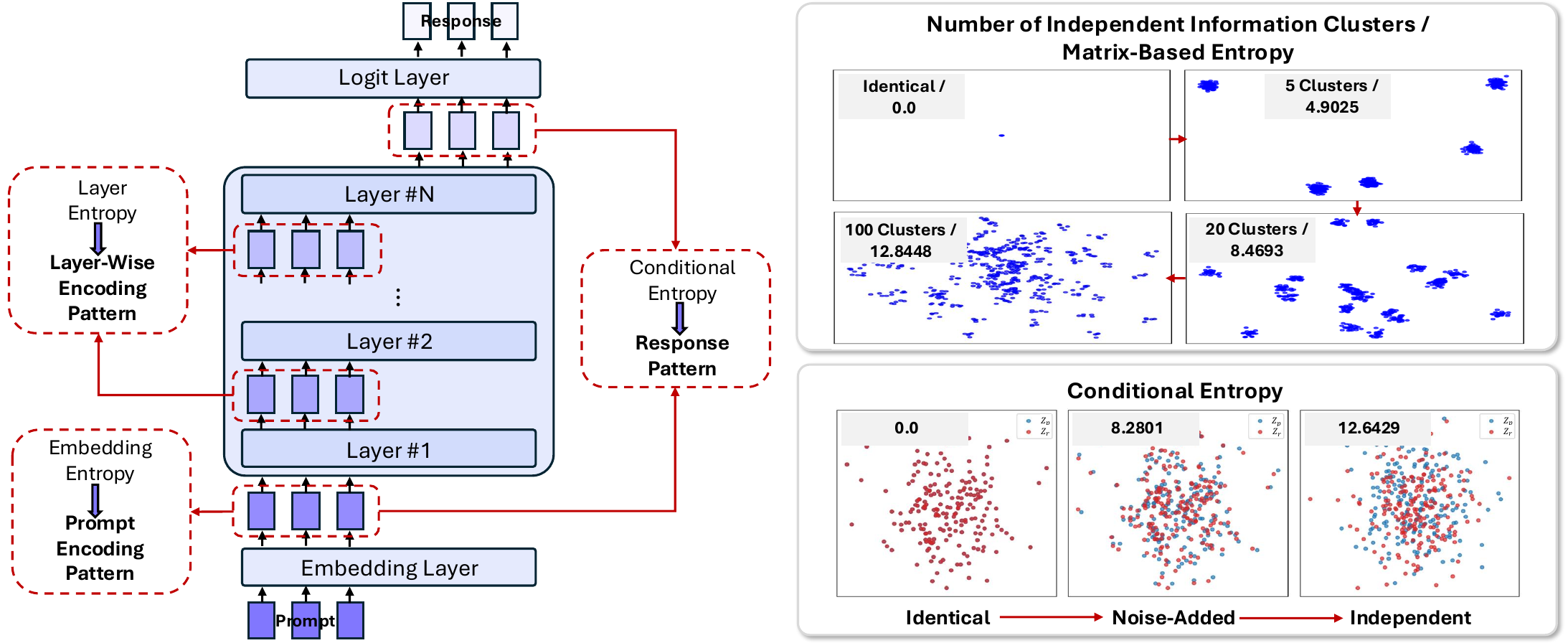}
    \vspace{-0.6cm}
    \caption{\textbf{Information-Theoretic Probing of UMMs.} Left: Extract prompt, response, and hidden-state embedding sequences from a Transformer-based UMM and compute entropy (measuring representational quality for encoding patterns) and conditional entropy (measuring output uncertainty given the input for response patterns). Right-Top: Matrix-based entropy increases with the number of independent information clusters. Right-Bottom: Conditional entropy rises as the dependency between two embedding sequences decreases.}
    \label{fig:entropy}
    \vspace{-0.4cm}
\end{figure*}

\subsection{Modeling UMMs via an Implicit Joint Distribution}
\label{subsec:implicit_joint}
We formalize a UMM as learning an implicit joint probability distribution $P(\mathbf{X}, \mathbf{Y})$ over visual inputs $\mathbf{X}$ (image patch sequences) and text inputs $\mathbf{Y}$ (text token sequences). Under this view, multimodal tasks correspond to conditional operations on this shared distribution. For example, image captioning implements $P(\mathbf{Y_{r}} \mid \mathbf{X} , \mathbf{Y_{p}})$, where $\mathbf{Y_p}$, $\mathbf{Y_r}$ denote prompt, response text, and text-to-image generation implements $P(\mathbf{X} \mid \mathbf{Y})$. The degree to which a UMM achieves genuine unification thus hinges on the internal coherence of this implicit joint model, which can be probed by entropy and condition entropy.

\subsection{Entropy and Conditional Entropy}
\label{subsec:info_theory}

The entropy $H(\mathbf{Z})$ of a embedding sequence $\mathbf{Z}$ reflects its uncertainty and effective dimensionality which higher entropy indicates a more isotropic and information-rich embedding space. More critically, let $\mathbf{Z_p}$ and $\mathbf{Z_r}$ denote as prompt embedding sequence and response embedding sequence, the conditional entropy $H(\mathbf{Z_{r}}\mid \mathbf{Z_{p}})$ captures the residual uncertainty in the output given the input, directly reflecting the model’s response behavior, which low values indicate fidelity-driven generation, and high values reflect creative responses. By comparing models’ entropy and conditional entropy across different prompts and responses, we can probe their representational and behavioral patterns across modalities and tasks, thereby analyzing the degree of unification.

\subsection{Challenges in Classic Entropy Estimation}
\label{subsec:challenges}
Despite its conceptual clarity, direct estimation of $H(\mathbf{Z})$ and $H(\mathbf{Z_{r}}\mid \mathbf{Z_{p}})$ is infeasible in practice. Transformer-based UMMs do not expose explicit probability densities. Instead, they produce high-dimensional, variable-length embedding sequences. Classical density-based estimators are unstable or undefined in such settings, necessitating a non-parametric alternative that operates solely on representational geometry.

\subsection{Matrix-Based Rényi Entropy}
\label{subsec:matrix_entropy}

We estimate entropy via matrix-based Rényi entropy~\citep{renyi1961measures}, which quantifies information content from the similarity structure of representations in kernel space~\citep{other-layer-by-layer,li2024large,skean2023dime}. Given a sequence of embeddings \( \mathbf{Z} = \{\mathbf{z}^{(i)}\}_{i=1}^n \in \mathbb{R}^d \), we first construct a Gram (kernel) matrix \( \mathbf{K} \in \mathbb{R}^{n \times n} \) using a Gaussian kernel:
\begin{equation}
    [\mathbf{K}]_{ij} = \exp\left( -\frac{\|\mathbf{z}^{(i)} - \mathbf{z}^{(j)}\|^2}{2\sigma^2} \right),
\end{equation}
where \( \sigma > 0 \) is a bandwidth parameter. Normalizing \( \mathbf{K} \) by its trace yields a valid probability matrix \( \mathbf{A} = \mathbf{K} / \mathrm{tr}(\mathbf{K}) \). The \( \alpha \)-order Rényi entropy of the representation is then defined as
\begin{equation}
    H_\alpha(\mathbf{K}) = \frac{1}{1 - \alpha} \log \left( \mathrm{tr}(\mathbf{A}^\alpha) \right).
\end{equation}
In practice, we set \( \alpha = 1.01 \) to approximate Shannon entropy while maintaining numerical stability. This formulation allows entropy to be computed directly from kernel matrices, enabling consistent estimation across heterogeneous modalities and variable-length sequences.

\noindent
\textbf{Validation of Entropy Sensitivity.} To validate that $H_\alpha(\mathbf{K})$ meaningfully reflects representational diversity, we conduct a controlled experiment in which we synthesize embedding sequences with varying numbers of independent information clusters. Specifically, we generate four sequences: all embeddings identical, embeddings sampled from 5 distinct Gaussian clusters, from 20 clusters, and from 100 clusters. As shown in the Right-Top of Fig.~\ref{fig:entropy}, the matrix-based entropy increases monotonically with the number of clusters, from zero for the uniform sequence to substantially higher values for the 100-cluster case. The corresponding 2D visualizations confirm that greater cluster separation correlates with higher entropy.

\subsection{A Proxy for Conditional Entropy}
\label{subsec:proxy}

To estimate conditional entropy without access to explicit densities, we leverage the identity
\begin{equation}
    H(\mathbf{Z_r} \mid \mathbf{Z_p}) = H(\mathbf{Z_p}, \mathbf{Z_r}) - H(\mathbf{Z_p}).
\end{equation}

Let \( \mathbf{Z_p} = \{\mathbf{z}_p^{(i)}\}_{i=1}^n \) and \( \mathbf{Z_r} = \{\mathbf{z}_r^{(i)}\}_{i=1}^m \) denote the prompt and response embedding sequences, respectively. We compute the input entropy \( H_\alpha(\mathbf{Z_p}) \) from the self-kernel matrix \( \mathbf{K}_{pp} \), where
\begin{equation}
    [\mathbf{K}_{pp}]_{ij} = \exp\left( -\frac{\|\mathbf{z}_p^{(i)} - \mathbf{z}_p^{(j)}\|^2}{2\sigma^2} \right).
\end{equation}

For the joint entropy \( H(\mathbf{Z_p}, \mathbf{Z_r}) \), we construct the block joint kernel matrix
\begin{equation}
    \mathbf{K}_{joint} = 
    \begin{bmatrix}
        \mathbf{K}_{pp} & \mathbf{K}_{pr} \\
        \mathbf{K}_{rp} & \mathbf{K}_{rr}
    \end{bmatrix},
\end{equation}
with cross-kernel entries defined as
\begin{equation}
    [\mathbf{K}_{pr}]_{ij} = \exp\left( -\frac{\|\mathbf{z}_p^{(i)} - \mathbf{z}_r^{(j)}\|^2}{2\sigma^2} \right),
\end{equation}
and \( \mathbf{K}_{rp} \) defined analogously. The joint entropy is then estimated as the matrix-based Rényi entropy of \( \mathbf{K}_{joint} \), denoted \( H_\alpha(\mathbf{K}_{joint}) \).

Our conditional entropy proxy is defined as
\begin{equation}
    \widehat{H}(\mathbf{Z_r} \mid \mathbf{Z_p}) := H_\alpha(\mathbf{K}_{joint}) - H_\alpha(\mathbf{K}_{pp}).
    \label{eq:cond_entropy_proxy}
\end{equation}

\noindent
\textbf{Further Interpretation and Validation.} This proxy quantifies the residual uncertainty in the response given the prompt, captured as the additional structural complexity in $\mathbf{K}_{joint}$ beyond $\mathbf{K}_{pp}$. To validate its sensitivity to semantic dependence, we conduct a controlled experiment: starting from a base embedding sequence $\mathbf{Z_p}$, we construct three response sequences, which are $\mathbf{Z_r} = \mathbf{Z_p}$ (perfect alignment), $\mathbf{Z_r}$ with mild Gaussian perturbations (partial alignment), and $\mathbf{Z_r}$ sampled independently (no alignment). As shown in the Right-Bottom of Fig.~\ref{fig:entropy}, $\widehat{H}(\mathbf{Z_r} \mid \mathbf{Z_p})$ increases monotonically as the dependency between $\mathbf{Z_p}$ and $\mathbf{Z_r}$ weakens: near-zero for identical sequences, moderate for perturbed ones, and highest for unrelated pairs. This confirms that the proxy behaves as expected, which lower values indicate strong input–output dependency (high fidelity), while higher values reflect greater randomness (high creativity). Although this proxy is not a formal Shannon conditional entropy, its alignment with semantic dependency makes it a theoretically grounded and empirically reliable measure for diagnosing pseudo-unification in UMMs.

%% file: sec/4_probing_framework_setting.tex
\section{Probing Framework and Setting}
\label{sec:framework}

\begin{figure*}[t]
    \centering
    \includegraphics[width=\linewidth]{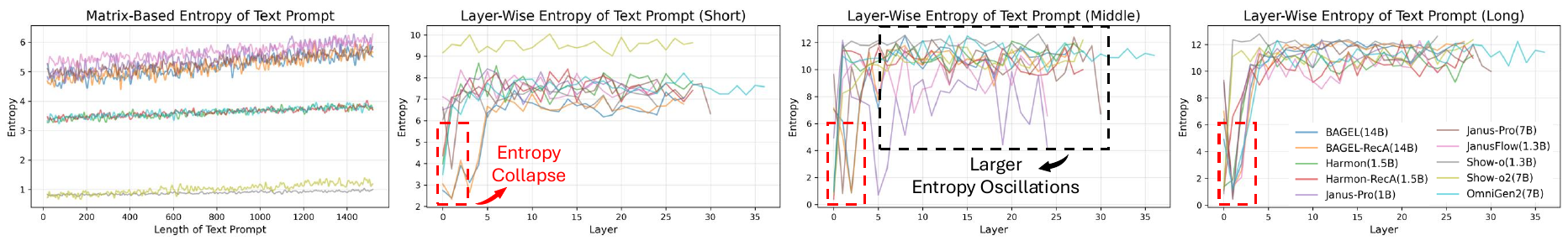}
    \vspace{-0.8cm}
    \caption{\textbf{Effect of Text Prompt Length on Embedding Entropy (1st) and Layer Entropy (2nd-4th).} 1st Sub-Fig: Entropy of text prompts increases with length, but absolute levels vary by architecture. 2nd-4th Sub-Figs: UMMs exhibit scale- and architecture-dependent early-layer compression strategies (\textit{e.g.}, entropy collapse), and middle-length prompts uniquely show larger entropy oscillations.}
    \label{fig:input_text_len}
    \vspace{-0.2cm}
\end{figure*}

\begin{table*}[t]
\scriptsize
\centering
\caption{\textbf{Embedding Entropy Results across Different Prompt Types.} The key distinction lies in entropy differences across modalities rather than variations within prompt types of the same modality.}
\vspace{-0.3cm}
\label{tab:probing_results}

\begin{tabular}{ccccccccccc}
\toprule
\textbf{Model} & \multicolumn{4}{c}{\textbf{Text Prompt Types~\citep{T2I-CoReBench}}} & \multicolumn{6}{c}{\textbf{Image Prompt Types~\citep{MMBench}}} \\
\cmidrule(r){2-5} \cmidrule(l){6-11}
& Composition & Abductive & Inductive & Deductive & Attribute & Logical & Relation & Coarse & Single & Cross \\
\midrule
BAGEL (14B)~\citep{bagel}        & 6.3884 & 6.4409 & 6.4741 & 5.8281 & 9.7289 & 9.6304 & 9.9517 & 10.0747 & 9.8128 & 9.8001 \\
BAGEL-RecA (14B)~\citep{bagel,reca}    & 6.2049 & 6.4110 & 6.4508 & 5.8298 & 9.6997 & 9.6470 & 10.0021 & 9.7920 & 9.8864 & 9.9293 \\
Harmon (1.5B)~\citep{harmon}       & 4.3885 & 4.3161 & 4.2649 & 4.1422 & 3.8095 & 5.7185 & 4.4754 & 4.9831 & 5.2457 & 3.6590 \\
Harmon-RecA (1.5B)~\citep{harmon,reca}  & 4.3669 & 4.2476 & 4.2127 & 4.1348 & 4.7023 & 5.3734 & 4.7539 & 4.7497 & 3.6590 & 4.8509 \\
Janus-Pro (1B)~\citep{janus-pro}      & 6.1642 & 6.8678 & 6.6165 & 6.1928 & 8.9038 & 8.2558 & 8.8812 & 7.3734 & 8.6797 & 8.8644 \\
Janus-Pro (7B)~\cite{janus-pro}      & 6.4311 & 6.8196 & 6.6549 & 5.3281 & 9.4128 & 9.7656 & 9.3537 & 8.7254 & 9.5013 & 8.2547 \\
JanusFlow (1.3B)~\cite{janusflow}    & 6.7410 & 6.0598 & 7.0411 & 6.4805 & 9.4408 & 9.2366 & 9.4352 & 8.5280 & 9.2743 & 9.3522 \\
Show-o (1.3B)~\citep{showo}       & 1.1955 & 1.2051 & 1.1650 & 1.1467 & 9.1688 & 9.1676 & 9.1695 & 9.1229 & 9.1692 & 9.1415 \\
Show-o2 (7B)~\citep{show-o2}        & 1.0548 & 1.0039 & 1.3433 & 0.9679 & 9.4913 & 9.4826 & 9.4027 & 9.2815 & 9.0923 & 9.3108 \\
OmniGen2 (7B)~\citep{omnigen2}       & 3.9031 & 3.8819 & 4.0090 & 3.7270 & 5.9166 & 7.7837 & 8.0143 & 7.2077 & 7.1494 & 7.3206 \\
\bottomrule
\end{tabular}
\vspace{-0.3cm}
\end{table*}

\subsection{Two-Level Probing Framework}
Building on the formulation in Sec.~\ref{sec:entropy_formulation}, we propose a two-level probing framework to diagnose pseudo-unification in UMMs. As shown in Fig.~\ref{fig:entropy}, at the prompt level, we analyze prompt representations by computing prompt entropy and its layer-wise entropy for text and image inputs. This reveals how each modality encodes semantic information and exposes asymmetries in representational geometry across modalities. At the response level, we trace prompt–response dependencies by estimating our conditional entropy proxy across layers. This allows us to identify whether a model exhibits divergent response patterns.

\subsection{Model Selection} 

We evaluate ten state-of-the-art UMMs: BAGEL (14B)~\citep{bagel} and Bagel-RecA (14B)~\citep{bagel,reca}: MoT-based models using Flow Matching~\cite{esser2024scaling} for image generation; Harmon (1.5B)~\citep{harmon} and Harmon-RecA (1.5B)~\citep{harmon,reca}: lightweight all-in-one models employing a Masked Autoencoder (MAE)~\cite{he2022masked} for vision; Janus-Pro (7B)~\citep{janus-pro} and Janus-Pro (1B)~\citep{janus-pro}: all-in-one architectures based on VQ-VAE~\citep{van2017neural} tokenization; JanusFlow (1.3B)~\citep{janusflow}: an all-in-one model using Flow Matching; Show-o (1.3B)~\citep{showo} and Show-o2 (7B)~\citep{show-o2}: all-in-one frameworks built on Diffusion Loss~\citep{yang2023diffusion} and Flow Matching, respectively; OmniGen2 (7B)~\citep{omnigen2}: a multimodal LLM integrated with a diffusion-based image generator. This selection encompasses three key dimensions of variation: (i) Architecture: all-in-one vs. MOT vs. two-stage; (ii) Image Generation Paradigm: Diffusion Loss, Flow Matching, VQ-VAE, and MAE; (iii) Model Scale: from 1B to 14B parameters. Moreover, by including RecA~\citep{reca} variants, we further probe the impact of post-training refinements on unification.

\subsection{Data Source: Text and Image Prompts}

Our probing experiments are conducted on two established multimodal benchmarks. For text prompts, we adopt T2I-CoReBench~\citep{li2025easier}, which comprises 1,080 prompts spanning composition and three types of reasoning tasks (deductive, inductive, and abductive), with lengths ranging from tens to approximately 1,500 characters. For image prompts, we use MMBench~\citep{MMBench}, containing 3,217 images covering both reasoning (attribute, logical, and relational) and perception (coarse, single-instance, and cross-instance).

%% file: sec/5_prompt.tex
\section{Prompt Representation}
\label{sec:prompt}

\begin{tcolorbox}[width=\columnwidth]
\small
\textbf{Key Insight:} \textit{Across text and image modalities, UMMs exhibit structure-agnostic encoding, which embedding entropy and layer-wise entropy dynamics are shaped by architecture and scale rather than semantic structure.}
\end{tcolorbox}

\subsection{Text Prompt}

\subsubsection{Effect of Prompt Length on Embedding Entropy}
The effect of length on prompt entropy is in 1st Sub-Fig of Fig.~\ref{fig:input_text_len}: (i) Entropy increases monotonically with prompt length, reflecting that longer prompts activate more independent directions in representation space, yielding richer and more isotropic embeddings. (ii) Absolute entropy levels, vary significantly by architecture. Models sharing the same LLM backbone exhibit similar baselines, highlighting the LLM prior’s dominant role in shaping representational geometry. Moreover, stronger LLMs retain higher isotropy after fine-tuning, indicating greater resilience to representational collapse under cross-modal alignment pressure.

\subsubsection{Effect of Prompt Length on Layer Entropy}

We analyze layer-wise entropy trajectories across text lengths in 2nd-4th Sub-Figs of Fig.~\ref{fig:input_text_len}: (i) Model-Dependent Early Compression: Deep layers usually retain high entropy, but early-layer behavior differs. Large models show early \textbf{\textit{entropy collapse}}, implying aggressive initial compression that favors cross-modal alignment over textual detail. Smaller models display smoother, oscillatory entropy growth, preserving input information. Show-o2 (7B) is atypical: it reaches high entropy quickly on short prompts but delays entropy increase for longer prompts, suggesting limited early scaling of representational capacity. (ii) Instability of Middle-Length Prompts: Middle prompts show \textbf{\textit{larger entropy oscillations}} in deep layers than both short and long prompts. We hypothesize they occupy an ``alignment ambiguity zone'', which too long for local context modeling, yet too short to activate hierarchical reasoning.

\begin{figure}[t]
    \centering
    \includegraphics[width=\linewidth]{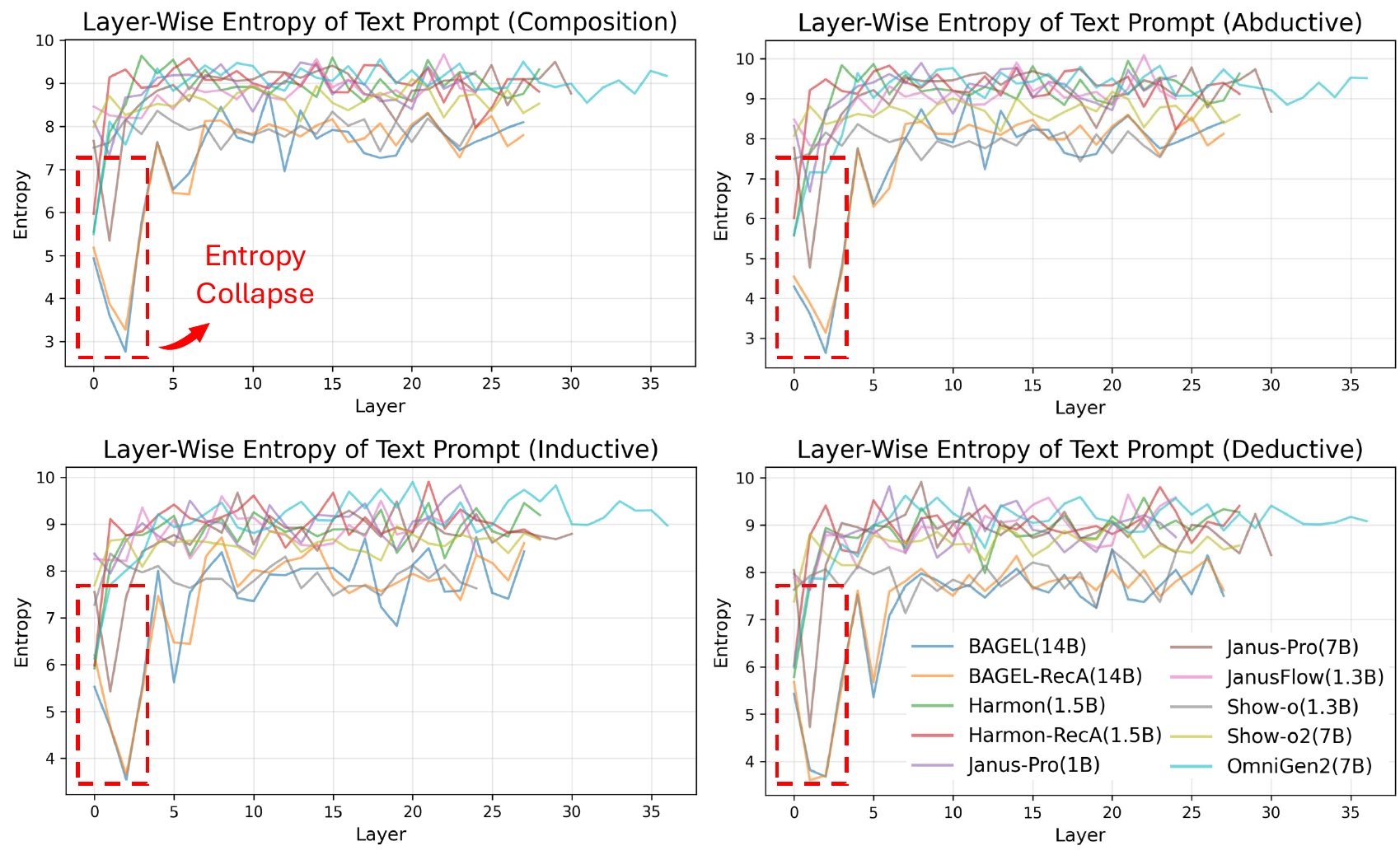}
    \vspace{-0.7cm}
    \caption{\textbf{Effect of Text Prompt Type on Layer Entropy.} The same model exhibits nearly identical layer-wise entropy dynamics across different text types.}
    \label{fig:input_text_type}
    \vspace{-0.5cm}
\end{figure}

\subsubsection{Effect of Prompt Type on 
Layer Entropy}

As shown in Tab.~\ref{tab:probing_results}, embedding entropy levels across text types are similar. As shown in Fig.~\ref{fig:input_text_type}, their layer-wise entropy trajectories are nearly identical across types. (i) Structure-Agnostic Encoding: UMMs follow a structure-agnostic process where reasoning cues are not preserved, and prompt engineering mainly elicits surface-level pattern matching rather than logical differentiation. (ii) Scale-Dependent Dynamics: Model size strongly shapes encoding behavior. Large models show early entropy collapse due to aggressive compression, while smaller ones maintain higher deep-layer entropy, preserving more semantic diversity. This challenges the belief that scaling alone improves representational fidelity and underscores the need for architectural innovation beyond mere parameter growth.

\subsection{Image Prompt}

We probe image prompt representations across perception-based and reasoning-based tasks, spanning low to high semantic density and structural complexity. Despite this diversity, all models exhibit nearly identical layer-wise entropy trajectories across image types, revealing that UMMs encode visual inputs in a structure-agnostic manner, governed by architectural priors rather than semantic or cognitive demands. Three distinct model-level patterns emerge:
(i) Harmon (1.5B) shows gradual entropy growth, reflecting conservative, detail-preserving encoding;
(ii) OmniGen2 (7B) maintains moderate, stable entropy, consistent with its decoupled MLLM+diffusion design and limited cross-modal interaction;
(iii) Mainstream UMMs (\textit{e.g.}, BAGEL, Janus, Show-o) exhibit immediate high-entropy saturation, suggesting early construction of a shared semantic space at the expense of task-sensitive processing.

\begin{figure}[t]
    \centering
    \includegraphics[width=\linewidth]{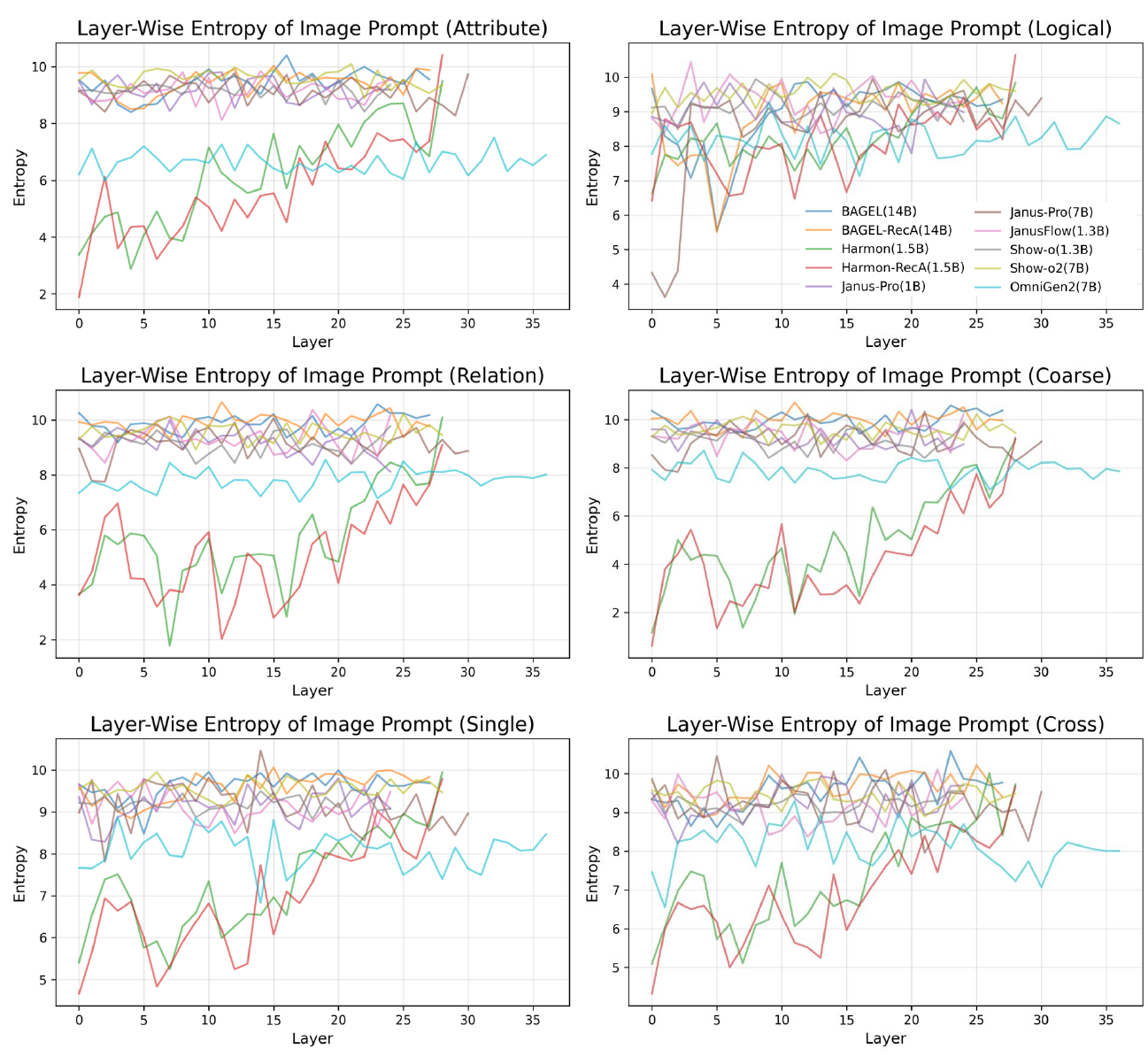}
    \vspace{-0.8cm}
    \caption{\textbf{Effect of Image Prompt Type on Layer Entropy.} The same model exhibits nearly identical layer-wise entropy trajectories across image types.}
    \label{fig:input_image_type}
    \vspace{-0.5cm}
\end{figure}

\subsection{Pseudo-Unification in Prompt Encoding}

\begin{tcolorbox}[width=\columnwidth]
\small
\textbf{Key Insight:} \textit{Despite architectural unification, all UMMs exhibit systematic cross-modal asymmetries in prompt encoding, differing in initial entropy and convergence dynamics, which reveals a heterogeneous representation space that underlies divergent generative behaviors and reinforces pseudo-unification.}
\end{tcolorbox}

After analyzing layer-wise representations across models, we further examine how \textbf{\textit{each individual UMM}} processes text and image prompts differently within its own architecture. Despite sharing a common backbone, all models exhibit systematic asymmetries between modalities, revealing that the ``unified'' representation space remains heterogeneous. Our layer-wise entropy comparisons yield four distinct patterns:

\begin{figure*}[t]
    \centering
    \includegraphics[width=\linewidth]{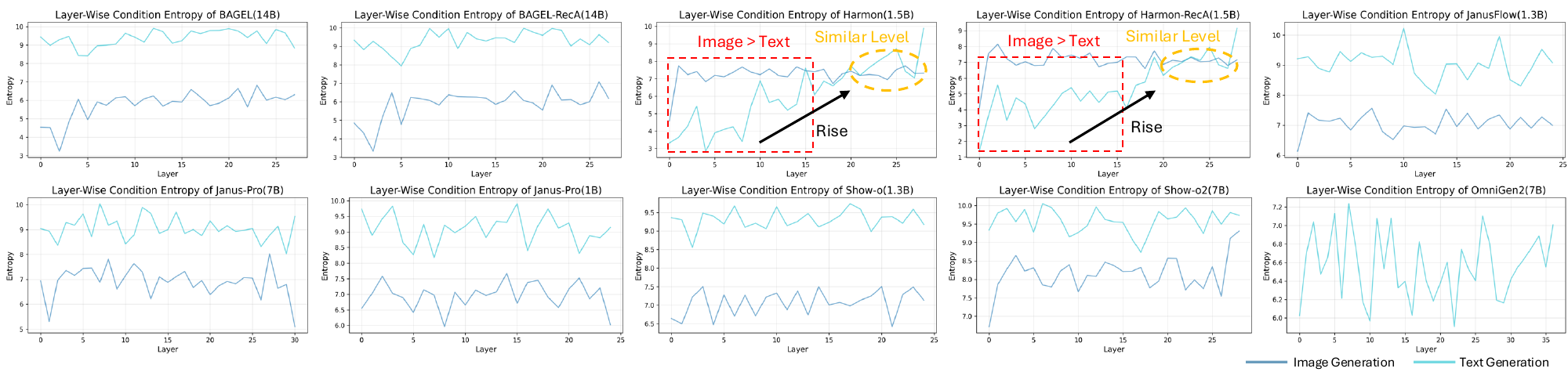}
    \vspace{-0.8cm}
    \caption{\textbf{Response Patterns of Different UMMs.} Except for Harmon, all UMMs exhibit a divergent response pattern, where layer-wise text conditional entropy consistently exceeds that of images. In contrast, Harmon shows a unique cross-modal convergence, with conditional entropy aligning to a similar level in the final layers. For Omnigen2, the image response directly uses the prompt-encoding layer, making prompt and response embeddings identical. Thus, layer-by-layer conditional entropy is omitted from the figure.}
    \label{fig:condition}
    \vspace{-0.4cm}
\end{figure*}

\begin{itemize}
    \item BAGEL series: Text prompts suffer from early-layer entropy collapse, rebounding only later to a moderate plateau of $5\sim6$. Image prompts, however, begin at a high entropy ($\approx9$) and remain stable throughout. This stark contrast (\textit{i.e.}, low linguistic entropy versus high visual entropy) implies that cross-modal alignment disproportionately compresses text representations while granting visual pathways greater representational freedom.
    \item Harmon series: Text prompts show a rapid entropy rise in early layers, stabilizing with mild oscillations around $7\sim8$. In contrast, image prompts start from a lower baseline and ascend gradually to the same range. This asynchronous convergence suggests a modality-asymmetric encoding strategy: language is encoded aggressively, while vision is built up conservatively.
    \item Show-o and Janus series: Despite using different image-generation objectives (\textit{e.g.}, diffusion and flow matching), both families display consistent cross-modal behavior: entropy for both text and image prompts surges early and plateaus. Differences lie only in absolute levels (\textit{e.g.}, text$\approx8$, image$\approx9$), indicating a design bias toward rapid saturation into a high-dimensional shared space, though without eliminating modality-specific scale offsets.
    \item OmniGen2: Though not a native UMM, it exhibits a hybrid pattern. Text entropy rises slowly before stabilizing at a high level in middle-to-deep layers, while image entropy starts high from layer 0 and remains flat. Notably, the final entropy values for both modalities are nearly identical, possibly due to its reliance on text-derived conditions that implicitly calibrate visual representations.
\end{itemize}

Collectively, these findings reveal a critical insight: while models treat semantic variations within a modality in a structure-agnostic manner, they consistently differentiate between modalities in their encoding dynamics. This cross-modal representational misalignment (\textit{i.e.}, evident in initial entropy levels, convergence speed, and stable states) likely underlies the divergent response patterns observed downstream. When vision and language follow distinct geometric trajectories from the onset, their generative behaviors cannot easily conform to a shared reasoning logic, thereby reinforcing pseudo-unification at the behavioral level.

%% file: sec/6_response.tex
\section{Response Pattern}
\label{sec:experiment}

After probing prompt representations, we find that differences in encoding patterns are not driven by prompt type, but are instead model-specific. The variation manifests as systematic discrepancies in how each model internally processes prompts from different modalities. Therefore, in our subsequent analysis of response patterns, we directly compare how each model generates text versus images.

\begin{tcolorbox}[width=\columnwidth]
\small
\textbf{Key Insight:} \textit{Nearly all UMMs exhibit a ``creative text vs. fidelity image'' response divergence, which is the evidence of pseudo-unification driven by misaligned generative patterns. Real unification requires shared inductive biases, not just shared parameters.}
\end{tcolorbox}

\subsection{Pseudo-Unification in Response Pattern}

We conduct a layer-wise comparison of conditional entropy (\textit{i.e.}, the uncertainty of the response given the prompt) between text generation and image generation. The results reveal a striking pattern: Nearly all UMMs exhibit significant cross-modal inconsistency in their response patterns.

Specifically, except for Harmon, every model exhibits higher conditional entropy in text generation than in image generation. This divergence reflects a fundamental difference in generative paradigms: (i) The higher entropy in text generation corresponds to a creative response pattern: after understanding the prompt, the model samples from a broad semantic distribution to produce plausible outputs (\textit{e.g.}, open-ended answers or narratives), embracing diversity and creativity. (ii) The lower entropy in image generation reflects a fidelity response pattern: the model aims to produce a single, highly deterministic visual output that strictly aligns with the prompt, suppressing randomness to ensure faithfulness. This split confirms that current UMMs do not unify generative logic. Instead, they inherit modality-specific optimization objectives, which preserve the open, probabilistic nature of LLMs for text, while adopt the fidelity constraints of diffusion or autoregressive vision models for images. This results in a ``dual-track'' response mechanism, a hallmark of pseudo-unification.

\subsection{More Discussion on Harmon}

Harmon stands out as the only model where image-generation conditional entropy exceeds that of text generation in early layers, with text entropy steadily rising across depth and eventually surpassing image entropy in the final layers. This reflects a fundamentally unified generative logic rooted in its architecture: Harmon uses a masked autoencoder for images, paralleling next-token prediction in text. Both modalities share the same inductive bias, which is contextual prediction: masked patches from visible context in vision, future tokens from prior context in language. Harmon thus provides evidence that pseudo-unification arises may from misaligned generative paradigms. Real unification may require grounding all modalities in a common contextual-prediction framework, treating visual and textual generation as structured inference from partial observations.

%% file: sec/7_conclusion.tex
\section{Conclusion and Future Direction}
\label{sec:conclusion}

Our work reveals that current UMMs suffer from \textbf{\textit{pseudo-unification}}: despite sharing parameters, they exhibit divergent representation and response patterns across modalities. This disconnect stems not from insufficient capacity (\eg, bigger model size and better backbone), but from misaligned generative inductive biases: text retains LLM-like creativity, while vision adheres to fidelity. The rare exception, Harmon, demonstrates that real unification is possible when both modalities are grounded in a shared contextual prediction paradigm. These findings urge the community to refocus on the original motivation for UMMs: synergy, not just multi-yet-decoupled task performance. Merely scaling architectures or curating more benchmarks will not suffice if the underlying information flow remains fragmented. 

\noindent
\textbf{Future Direction.} Some works~\citep{kang2025rare,liu2025visual,zhang2025memory} explore adjusting the prompt input to reshape entropy and thus enhance context understanding, but we found it difficult to change the information patterns already learned by the model from the prompt. Further works are worth further probing and designing for information consistency, including (i) Rethinking pre-training objectives to enforce unified entropy dynamics, such as through symmetric prediction tasks across modalities; (ii) Moving beyond ``does it work?'' evaluations toward ``how and why does it unify?'' analyses. Only by treating information patterns as first-class design criteria, not emergent byproducts, can we move from pseudo-unification to genuine multimodal synergy.